\documentclass[runningheads]{llncs}

\usepackage{eccv}

\usepackage{eccvabbrv}

\usepackage{graphicx}
\usepackage{booktabs}
\usepackage{multirow}
\usepackage{wrapfig}
\usepackage{pifont} 
\usepackage{makecell}
\newcommand{\cmark}{\ding{51}}
\newcommand{\xmark}{\ding{55}}

\usepackage[accsupp]{axessibility}  

\usepackage{hyperref}

\usepackage{orcidlink}

\begin{document}

\title{DeWorldSG: Depth-Aware 3D Semantic Scene Graph Generation via World-Model Priors} 

\titlerunning{DeWorldSG}

\author{Seok-Young Kim\inst{1}\orcidlink{0009-0008-7336-5699} \and
Abdelrahman Elskhawy\inst{2,3}\orcidlink{0009-0004-2429-7539} \and
Taewook Ha\inst{1}\orcidlink{0009-0004-7946-8577} \and
Dooyoung Kim\inst{4}\orcidlink{0000-0002-6003-2181} \and
Eunjae Shin\inst{1}\orcidlink{0009-0005-9297-3487} \and
Benjamin Busam\inst{2,3}$^{\dagger}$\orcidlink{0000-0002-0620-5774} \and
Woontack Woo\inst{1}$^{\dagger}$\orcidlink{0000-0002-5501-4421}}

\authorrunning{SY.Kim et al.}

\institute{Korea Advanced Institute of Science and Technology  
\\
\and
Technical University of Munich\\
\and
Munich Center for Machine Learning (MCML)\\
\and
La Trobe University \\
\email{seokyoung@kaist.ac.kr}}

\maketitle

\begingroup
\renewcommand{\thefootnote}{\fnsymbol{footnote}}
\footnotetext{Co-corresponding authors.}
\endgroup

\vspace{-1.0 em}
\begin{center}
\small
\href{https://deworldsg2026.github.io}
{\textcolor{black}{\texttt{https://deworldsg2026.github.io}}}
\end{center}

\begin{abstract}
We present \textit{DeWorldSG}, a novel framework that generates spatio-temporally robust 3D Semantic Scene Graphs from RGB-D sequences. Existing methods often struggle to construct reliable 3D scene graphs due to unstable 3D object representations and missing relations caused by frame-wise inference. DeWorldSG addresses these issues by estimating instance-level geometric 3D Gaussian distributions through depth-guided filtering and representing each object as a probabilistic 3D node rather than a single projected point. To mitigate relational sparsity from frame-wise inference, our framework further aggregates spatiotemporal evidence across object pairs and refines relations using contextual priors derived from a world model (V-JEPA 2). Experiments on the 3DSSG and ReplicaSSG datasets demonstrate state-of-the-art (SoTA) performance in both object and predicate prediction, while producing temporally consistent scene structures. In particular, our method improves triplet recall by 77.4\% and predicate recall by 23.2\% over prior SoTA approaches, making it suitable for robotic manipulation and AR applications. Our code and models are open-sourced. 


\end{abstract}


\section{Introduction}
\label{sec:intro}

For robotic intelligence and Augmented Reality (AR) systems to operate reliably in real-world environments, it is necessary to go beyond recognizing individual objects and instead reason about the complex spatial relationships among them. To fundamentally perceive 3D space and perform high-level tasks, a compact structural representation that captures scene entities and their mutual relations is essential \cite{hydra, hydra2}. 3D Semantic Scene Graph (SSG) generation \cite{3dssg} addresses this need by transforming complex 3D environments into a compact graph structure consisting of object-level nodes and relational edges. Such a structural representation preserves both geometric information and semantic context of the scene, thereby providing a fundamental basis for higher-level cognitive tasks such as agent planning \cite{spotlight, intro2, intro3}, 3D content synthesis \cite{common, echo, lr, mmg, scenelinker}, and object augmentation \cite{imaginatear, intro6, rc} in AR. However, current 3D SSG generation approaches still suffer from unstable 3D geometry and inconsistent relational inference across video frames, often resulting in missing or unreliable relations in the final graph.

Recently, with the goal of enabling these interactive applications, increasing attention has been given to approaches \cite{kim, vgfm, fross} that first extract 2D scene graphs from RGB(-D) sequences and then progressively lift and merge them into a global 3D scene graph. This strategy reduces dependency on full 3D reconstruction, maintains robustness to geometric noise, and lessens sensitivity to camera pose errors, making it suitable for stable operation in real-world settings. Despite these advantages, unstable depth estimates near object boundaries can distort lifted object geometry and destabilize object-level 3D representations, leading to erroneous merging of distinct objects or fragmentation of a single one \cite{fross}. Moreover, because existing approaches \cite{monossg, sgfusion, fross, kim, vgfm, imp} infer relationships independently for each frame, limited observations caused by occlusion or low-resolution input often result in missing essential edges. As a result, the generated scene graphs tend to reflect local geometric proximity rather than coherent scene-level context.

\begin{figure*} [t]
  \includegraphics[width=\textwidth]{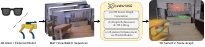}
  \caption{DeWorldSG generates a spatio-temporally robust 3D semantic scene graph from RGB-D observations. It combines 2D scene graph generation, depth-aware 3D lifting, and world-model-based relation enhancement to produce stable object nodes and context-aware relational edges for AR and embodied robotic systems.}
  \label{fig:teaser}
\end{figure*}

The root cause of these issues is treating scene graph generation as a sequence of independent frame-level predictions. In real-world environments, object geometry must be modeled with uncertainty during 2D-to-3D lifting, and semantic relationships should be inferred by accumulating evidence over time rather than from single observations. This calls for an incremental 3D SSG framework that maintains stable object representations while refining inter-object relations using spatiotemporal context.

In this work, we propose \textit{DeWorldSG}, a novel framework for context-grounded 3D SSG generation that combines depth-aware probabilistic object modeling with spatiotemporal relation reasoning (Fig.~\ref{fig:teaser}). Our approach constructs per-frame 2D scene graphs from RGB-D sequences and incrementally builds a global 3D graph while explicitly accounting for uncertainty in lifted object geometry. To enhance relational completeness and consistency, it aggregates relation evidence across temporally dispersed observations and incorporates video-based priors from a world model \cite{vjepa2} when local visual evidence is insufficient. Finally, confidence-based probabilistic fusion balances the influence of these priors against direct visual and geometric cues, suppressing hallucinated relations and yielding a more globally consistent 3D scene graph.

We evaluate DeWorldSG on the 3DSSG \cite{3dssg} and ReplicaSSG \cite{fross} datasets, which provide object class and relation annotations for indoor environments. Our method consistently outperforms existing baselines in both object classification and relationship prediction, achieving state-of-the-art performance across all metrics. Ablation studies further validate each component's contribution and demonstrate robust graph generation even without ground-truth camera poses, while sustaining incremental graph updates at interactive rates. These results indicate that DeWorldSG provides a reliable structural representation with high-fidelity relational reasoning, which is essential for downstream spatial reasoning applications such as robotic manipulation \cite{intro4, intro5} and context-aware AR anchoring \cite{scenelinker}.

\smallskip
\noindent The main contributions of this work are summarized as follows:
\begin{enumerate}
    \item We present \textit{DeWorldSG}, a novel framework for depth-aware 3D semantic scene graph generation from RGB-D sequences that represents each object as an instance-level probabilistic 3D Gaussian distribution derived from mask-guided depth observations.
    
    \item We propose Dual-Domain Depth Refinement (DR), a sequential filtering mechanism that effectively removes spatial and distributional depth noise, demonstrating robust instance-level 3D Gaussian estimation and stabilizing global 3D scene graph construction.

    \item We introduce a spatiotemporal relation reasoning scheme that aggregates relational evidence across frames and leverages world model priors to contextualize local observations, enhancing relational completeness and consistency.
\end{enumerate}

\section{Related Work}

\subsection{Geometry-Explicit 3D SSG Generation}

3D Semantic Scene Graphs \cite{rw_3dssg, rw_3dssg2, 3dssg} represent a 3D environment as a structured graph, where nodes correspond to object instances and edges encode semantic and spatial relationships, providing a compact yet expressive abstraction for embodied AI \cite{embodied1, embodied3} and robotics reasoning~\cite{booker2024embodiedrag, gu2024conceptgraphs, saxena2024grapheqa}. Early 3D SSG generation methods~\cite{3dssg, pham2024tesgnn, monossg, sgfusion} follow a geometry-first paradigm, explicitly leveraging point-clouds or SLAM-estimated 3D structure. In \cite{3dssg}, a learned pipeline regresses a semantic scene graph directly from reconstructed indoor point-clouds, extracting features for object instances and pairwise union regions and refining them with a GCN \cite{gcn} to jointly predict objects and relations. Similarly, \cite{monossg, sgfusion} leverage Visual-SLAM \cite{rw_slam, orbslam3} to estimate point-clouds and 3D bounding boxes from RGB(-D) sequences, and then construct scene graphs from the reconstructed 3D structure using GNN-based reasoning \cite{gnn}. The main advantage of this line of work is that relations are grounded in explicit 3D geometry from the outset, which historically made it effective for spatial reasoning. However, graph quality is tightly coupled to reconstruction quality, making these methods sensitive to pose drift, incomplete geometry, and computational overhead. DeWorldSG mitigates this bottleneck by estimating instance-level probabilistic geometry directly from RGB-D observations, enabling stable object nodes without requiring dense reconstruction.

\subsection{2D-to-3D Lifted 3D SSG Generation}
Lifting-based 3D SSG methods first predict per-frame 2D entities and relations from RGB(-D) observations and then align them in a shared 3D coordinate frame. ~\cite{vgfm} generates 3D scene graphs from multi-view image sequences by first estimating object 3D locations and occupancies from matched 2D detections via multi-view geometry, followed by relation prediction. Also, ~\cite{kim} enables incremental construction and updating of a global 3D scene graph from sequential observations for intelligent-agent reasoning. 
More recently, FROSS \cite{fross} directly lifts 2D scene graphs to 3D and merges objects as 3D Gaussian distributions ~\cite{merge}, facilitating online 3D SSG generation. Unlike earlier lifting pipelines that rely primarily on geometric alignment, FROSS introduces probabilistic Gaussian representations for object merging, improving robustness during incremental graph construction.
However, because the 2D-to-3D lifting pipeline of FROSS does not explicitly model object geometry or depth distributions, the generated scene graphs often contain spatially inconsistent or missing relations, which limits fine-grained spatial reasoning. In contrast, DeWorldSG explicitly estimates mask-guided instance-level depth distributions from filtered RGB-D evidence, yielding geometrically grounded 3D Gaussians that support more reliable graph reasoning than weakly constrained lifting.

\subsection{World Models for 3D Scene Understanding}
A world model is a predictive model that (i) infers and completes a latent state of the environment from partial observations \cite{vjepa, vjepa2, vljepa} and (ii) forecasts the evolution of this state to support internal rollouts for planning~\cite{lecun2022path}. In this work, we focus on latent-state inference and completion, exploiting it as a contextual prior for 3D scene understanding.
Recent works exploit this ability for 3D scene understanding by recovering occluded structure with explicit 3D Gaussian world models \cite{hu2025dsg}, imagining plausible unseen views with controllable video world models \cite{yang2025mindjourney, dwm}, or learning geometry-aware cross-modal latent representations \cite{zhang2025concerto}.


While these approaches primarily target geometric reconstruction, view synthesis, or representation learning, we instead incorporate world-model priors into a scene graph structure for structured 3D spatial reasoning. In scene graph generation, relations between instances are often under-constrained by instantaneous visual evidence due to occlusion, noisy geometry, and viewpoint-dependent ambiguities. A world model learns predictive representations that complete unobserved aspects of the current state and encode regularities and commonsense constraints over object interactions beyond per-frame appearance~\cite{lecun2022path}. Building on this insight, DeWorldSG leverages the contextual understanding of a pretrained video world model to infer relational priors across clips and fuses these priors with geometric evidence to construct a spatio-temporally consistent 3D semantic scene graph.

\section{Preliminary}
\textbf{3D Semantic Scene Graph.}
A 3D Semantic Scene Graph (SSG) abstracts a scene into a directed graph $G=(V,E)$ composed of object nodes and inter-object relation edges.
Each node $v_i \in V$ represents the $i$-th object instance and is defined by its 3D geometric parameters (position, scale, and spatial uncertainty) and a semantic category label $c_i$.
Each directed edge $e_{i\rightarrow j} \in E$ encodes a spatial relation between a subject $v_i$ and an object $v_j$, where the predicate type $r_{i\rightarrow j}$ is selected from a predefined predicate set $R$ (e.g., \textit{attached to}, \textit{under}).
Depending on downstream scene-graph applications, nodes can be extended with additional attributes such as visual descriptors, text embeddings, or temporal information.
In this paper, we focus on the core components required for semantic graph reasoning: $(c_i, \mu_i^{3D}, \Sigma_i^{3D})$ and $r_{i\rightarrow j}$.
Accordingly, our goal is to construct a 3D SSG $G$ from an RGB-D sequence $\{(I_t, D_t)\}_{t=1}^{T}$, given camera intrinsics $K$, and per-frame camera poses $T_t=[R_t \mid t_t]$.

~\\
\textbf{Object Representation with 3D Gaussians.}
In 3D SSG, each object node represents an instance in the scene, which can be defined using various geometric representations such as a 3D bounding box, a point-cloud or mesh, or a probability distribution.
Inspired by \cite{fross}, we model the observation of object instance $i$ at frame $t$ as a 3D Gaussian $\mathcal{N}(\mu_{i,t}^{3D}, \Sigma_{i,t}^{3D})$. 
Compared with point-cloud or mesh-based representations, this probabilistic form summarizes an object's position and spatial extent using low-dimensional statistics, resulting in improved computational efficiency.
In contrast to rigid 3D bounding-box representations, it provides a more faithful approximation of the underlying object's geometric structure.
Furthermore, this representation enables concise integration of multi-view observations through iterative updates of the mean and covariance, while facilitating stable object association during global graph merging via distribution-based distance metrics.
Unlike prior 2D-to-3D lifting approaches that approximate object geometry from projected 2D bounding boxes, we directly estimate instance-level 3D Gaussian distributions from depth observations. This probabilistic formulation provides a robust spatial foundation for downstream relation reasoning and global graph construction.

\begin{figure*} [t]
  \includegraphics[width=\textwidth]{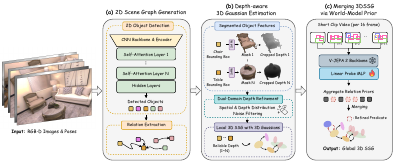}
  \caption{
    \textbf{Overview of DeWorldSG.} Given RGB-D image sequences, DeWorldSG first constructs frame-wise 2D scene graphs by detecting objects and extracting their relationships (a).
The detected instances are lifted to depth-aware probabilistic 3D Gaussians through mask-guided depth filtering and dual-domain depth refinement (b).
Finally, spatiotemporal predicate priors from a video world model are aggregated and fused with visual predictions to incrementally merge local graphs into a globally consistent 3D semantic scene graph (c).}
  \label{fig:overview}
\end{figure*}

\section{Method}

\subsubsection{Overview.}
We propose \textit{DeWorldSG}, a framework for generating spatially coherent 3D semantic scene graphs from RGB-D sequences.
As illustrated in Fig.~\ref{fig:overview}, the core of DeWorldSG is to represent each object as a probabilistic 3D Gaussian estimated from filtered instance-level depth observations.
For each frame, we first infer 2D object instances and pairwise relations from the input RGB-D sequence (Sec.~\ref{sec:4.1}).
We then refine the depth of each instance using SAM-guided masks and dual-domain filtering, and estimate a depth-aware 3D Gaussian that serves as the local object node.
These local graphs are incrementally merged into a global 3D scene graph based on semantic consistency and Gaussian similarity (Sec.~\ref{sec:4.2}).
Finally, ambiguous relational edges are refined by fusing temporally aggregated visual evidence with priors from a video world model through confidence-aware probabilistic fusion (Sec.~\ref{sec:4.3}).
This process yields stable, globally consistent 3D Semantic Scene Graphs that capture both geometric accuracy and context-aware relational structure.

\subsection{Depth-aware Local 3D Gaussian Representation}
\label{sec:4.1}
Relying on a single representative depth value for 2D-to-3D lifting inherently limits the stability of the approximated 3D Gaussian distribution.
In scenes with significant object overlap, boundary depth distortions and flying pixels introduce high variance into $\mu_{3D}$ and $\Sigma_{3D}$, often leading to fragmented or incorrectly merged instances in the global graph.
To overcome this limitation, \textit{DeWorldSG} reformulates object lifting as direct estimation of a 3D Gaussian distribution from instance-level depth observations, instead of inferring geometry from lifting 2D bounding boxes.

~\\
\textbf{2D Scene Graph Generation.}
For each frame $t$ of the input RGB-D sequence, we extract a 2D scene graph $G_t^{2D}$ from the RGB image $I_t$.
Following the strategy of \cite{fross}, we use an object detector \cite{detr, detr2} to obtain a per-frame instance set $\{o_{i,t}\}$, and predict a bounding box $b_{i,t}=(c_x,c_y,w,h)$ and class probability $p(c_{i,t})$ for each object.
A 2D relation inference model \cite{egtr} is then applied to each ordered object pair $(i,j)$ within the frame, producing relation logits $p(r_{i\rightarrow j,t})$.
As a result, the 2D scene graph at frame $t$ consists of nodes $\{(b_{i,t}, p(c_{i,t}))\}$ and edges $\{p(r_{i\rightarrow j,t})\}$, which are subsequently reprojected into 3D and serve as input for global graph construction.

~\\
\textbf{Box-Conditioned Instance Extraction.}
Given the estimated bounding box $b_{i,t}$, we generate an object-wise binary mask $M_{i,t}$ via semantic segmentation \cite{sam2}.
If masks from different classes assign the same pixel within a frame, we identify it as an ambiguous pixel and discard it from all masks to prevent depth mixing across objects.
Moreover, if the number of valid pixels in a mask is too small ($<10$ pixels), covariance estimation becomes numerically unstable. In this case, we deem the instance as an invalid observation and exclude it from subsequent steps.
This mask-based conditioning exploits object shape to produce cleaner depth samples for subsequent Gaussian estimation.

~\\
\textbf{Dual-Domain Depth Refinement.}
Despite the use of instance masks $M_{i,t}$, the corresponding depth map $D_t$ often contains outliers from multi-surface observations or overlapping objects, which bias the estimated 3D Gaussians.
We address this with Dual-Domain Refinement (DR), a sequential filtering algorithm operating in both \textit{spatial} and \textit{depth-distribution} domains.
In the \textit{spatial domain}, we enforce local depth consistency to preserve continuous geometric structures.
For each pixel $p \in M_{i,t}$, we compare its depth $D_t(p)$ to the median depth of its $3 \times 3$ neighborhood $\Omega(p)$.
Pixels that deviate significantly from the local median are filtered out as follows:
\begin{equation}
M_{\text{spatial}} =
\left\{\, p \in M_{i,t} \;\middle|\;
\left| D_t(p) - \tilde{D}_t(p) \right|
< \tau \, \tilde{D}_t(p) + \epsilon
\right\},
\label{eq:1}
\end{equation}
where 
$\tilde{D}_t(p) = \operatorname{median}_{q \in \Omega(p)} D_t(q)$,
and $\tau, \epsilon$ are tolerance hyperparameters.
This step suppresses discontinuous boundary noise (e.g., flying pixels) and yields a more coherent depth distribution around the object.

In the \textit{depth-distribution domain}, we further refine the spatial inliers via 1D clustering along the depth axis.
Let $Z_{\text{spatial}} = \{ D_t(p) \mid p \in M_{\text{spatial}} \}$ be the set of refined depth values, and define the reference depth as
$d_{\text{ref}} = \operatorname{median}(Z_{\text{spatial}})$.
To adapt to varying depth scales, the clustering threshold is dynamically determined as:

\begin{equation}
\Delta_{\text{thr}} =
\max\left(
\gamma_{\text{base}},
\alpha \cdot \operatorname{median}(\Delta z),
\beta \cdot d_{\text{ref}}
\right),
\label{eq:2}
\end{equation}
where $\Delta z$ denotes adjacent depth differences in the sorted depth set,
$\gamma_{\text{base}}$ enforces a minimum separation,
$\alpha$ scales the typical local depth variation,
and $\beta$ adjusts the threshold according to the global depth scale.
We then partition the sorted depth sequence using $\Delta_{\text{thr}}$, and retain the cluster whose mean depth is closest to $d_{\text{ref}}$ as the final inlier set $P_{\text{final}}$.
DR therefore isolates a dominant, self-consistent depth layer per instance, which is crucial for stable Gaussian parameter estimation and erroneous global merging prevention.

~\\
\textbf{Depth-guided 3D Gaussian Estimation.}
Given the refined inlier pixel set $P_{\text{final}} \subseteq M_{i,t}$, each pixel $(u,v) \in P_{\text{final}}$ is back-projected into 3D camera coordinates using depth $z = D_t(u,v)$, and then transformed into the world frame to form an instance-specific 3D point set $P_{i,t}$.
With camera intrinsics $K$, each pixel is mapped to a camera-coordinate point $\tilde{\mathbf{x}}$, and using the frame pose $T_t=[R_t\mid t_t]$ yields the world-coordinate point $\mathbf{x}_n$:
\begin{equation}
\tilde{\mathbf{x}} = z\,\mathbf{K}^{-1}
\begin{bmatrix}
u\\
v\\
1
\end{bmatrix},
\qquad
\mathbf{x}_n = \mathbf{R}_t \tilde{\mathbf{x}} + \mathbf{t}_t,
\ \mathbf{x}_n \in \mathbb{R}^3.
\label{eq:3}
\end{equation}
$P_{i,t}$ thus aggregates multiple refined depth observations, through mask-based instance extraction and DR, for instance $i$ in frame $t$. 
In contrast to representative-depth lifting, which relies on a single depth value per object, we explicitly model geometric uncertainty and surface spread by estimating a 3D Gaussian from $P_{i,t}$ via the sample mean and covariance:
\begin{equation}
\mu_{i,t}^{3D} =
\frac{1}{|P_{i,t}|}
\sum_{\mathbf{x}_n \in P_{i,t}} \mathbf{x}_n,
\quad
\Sigma_{i,t}^{3D} =
\frac{1}{|P_{i,t}|-1}
\sum_{\mathbf{x}_n \in P_{i,t}}
(\mathbf{x}_n-\mu_{i,t}^{3D})(\mathbf{x}_n-\mu_{i,t}^{3D})^{\top}
+ \epsilon \mathbf{I}.
\label{eq:4}
\end{equation}
where $\epsilon \mathbf{I}$ is a regularization term for numerical stability. 
Each object node is therefore represented as a probabilistic 3D Gaussian derived from instance-level depth-refined observations. Together with frame-wise relation predictions, these nodes form the local 3D scene graph $G_t^{3D} = (V_t, E_t)$ at frame $t$.

\subsection{Incremental Merging into the Global 3D SSG}
\label{sec:4.2}

DeWorldSG incrementally integrates each local 3D graph $G_t^{3D}=(V_t, E_t)$ into the global graph $G^{3D}$.
A key challenge is establishing cross-frame object identity between 3D Gaussian nodes and merging those that correspond to the same instance.
For each local node $v_i^t \in V_t$, we therefore search for a correspondence in the global node set $V^{3D}$.
To compare geometric similarity, we compute the Hellinger distance \cite{hell} between two Gaussians $\mathcal{N}(\mu_i,\Sigma_i)$ and $\mathcal{N}(\mu_j,\Sigma_j)$.
If this distance is below a threshold $\delta_g$, the nodes are considered to have compatible spatial location and dispersion structure.
In parallel, for class probability distributions $p_i, p_j \in \mathbb{R}^{|C|}$, we define an inner-product-based class distance:
\begin{equation}
D_c(i,j) =
1 - \sum_{k=1}^{|C|} p_i^{(k)} p_j^{(k)}.
\label{eq:5}
\end{equation}
This distance is small when two nodes assign high probability mass to the same class.
We merge nodes only when both geometric and semantic criteria are satisfied:
\begin{equation}
HD(i,j) < \delta_g
\quad \text{and} \quad
D_c(i,j) < \delta_c.
\label{eq:6}
\end{equation}

When these conditions hold, we fuse the two Gaussians using weighted averaging.
Let $w_i$ and $w_j$ denote their respective observation frequencies. The merged node $v_k$ is updated as:
\begin{equation}
\mu_k =
\frac{w_i \mu_i + w_j \mu_j}{w_i + w_j},
\qquad
\Sigma_k =
\frac{w_i \Sigma_i + w_j \Sigma_j}{w_i + w_j}
+
\frac{w_i w_j}{(w_i + w_j)^2}
(\mu_i - \mu_j)(\mu_i - \mu_j)^{\top}.
\label{eq:7}
\end{equation}
where ${\mu}_i$ and ${\mu}_j$ are the mean vectors, 
${\Sigma}_i$ and ${\Sigma}_j$ are the associated 
covariance matrices of the Gaussians being merged.

For relation edges, we reassign all edges connected to the merged nodes and aggregate the relation prediction distributions for each object pair.
The final relation is obtained by accumulating prediction probabilities across multiple frames and selecting the most probable relation.
This temporal aggregation mitigates noisy single frame predictions and yields a more stable and temporally consistent 3D scene graph.

\begin{figure*}[t]
\centering
\includegraphics[width=\linewidth]{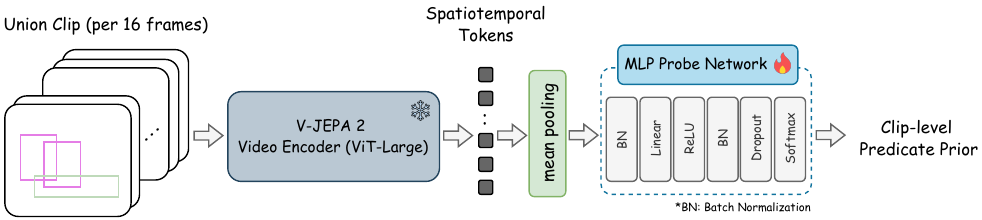}
\caption{
For each object pair, we extract a 16-frame union-crop clip and encode it with a frozen V-JEPA 2 backbone. The resulting spatiotemporal tokens are mean-pooled and passed to an MLP probe to produce a clip-level predicate distribution.
}
\label{fig:jepa_fig}
\end{figure*}

\subsection{3D Semantic Scene Graph Meets World Models}
\label{sec:4.3}
DeWorldSG is motivated by the premise that scene-graph relations should be refined through aggregated visual evidence, rather than inferred from a single-frame prediction.
To this end, we integrate geometric constraints from the global 3D scene graph with temporal visual context encoded by a world model.
In particular, V-JEPA~2 \cite{vjepa2} captures structural regularities and physical consistency beyond frame-level feature extraction.
We leverage it as a \textit{visual evidence collector} to refine relational edges in the global graph.

~\\
\textbf{Geometric Neighbor Pairing with Temporal Context.}
To reduce hallucinated relations, we do not treat all mask pairs within a frame as candidate edges.
Instead, we restrict inference to geometrically proximal object pairs.
Following \cite{prism}, we adopt a geometric proximity cost that combines the normalized 2D center distance $d_{2D}(i,j)$ and the depth disparity:
\begin{equation}
\mathrm{GeoCost}(i,j) =
0.5 \cdot \frac{d_{2D}(i,j)}{d_{\mathrm{diag}}}
+ |z_i - z_j|,
\qquad
\mathrm{GeoCost}(i,j) < 0.45.
\label{eq:8}
\end{equation}
In \cref{eq:8}, $d_{\mathrm{diag}}$ denotes the image diagonal length and $|z_i - z_j|$ is the absolute depth difference between the two objects.
Only pairs satisfying this constraint are retained.

For each selected pair $(i,j)$, we compute their union bounding box and extract the corresponding region from a sliding 16-frame buffer to construct a short video clip.
If one object becomes occluded or undetected, the clip is truncated at the last reliable observation.
When the valid interval is shorter than required, the most recent valid frame is repeated to maintain a fixed input length.
Using short temporal sequences instead of a single frame encourages predicates to stabilize based on recurring structural patterns rather than instantaneous geometric noise.

~\\
\textbf{Spatiotemporal Predicate Refinement with World-Model.}
From each 16-frame union-crop clip, we extract temporal representations that encode structural regularities and spatiotemporal context, which serve as evidence for predicate refinement.
We freeze the pretrained V-JEPA~2 backbone and attach a lightweight probe network consisting of two fully connected layers.
The probe is trained on the training split of each dataset to predict relation distributions.
Given the temporal representation, the MLP produces predicate probabilities 
$p_{\mathrm{WM}}(r_{i\rightarrow j})$ over predicate classes defined in the target dataset \cite{3dssg, fross}.

Rather than directly applying frame-wise predictions, we accumulate them at the class-pair level.
For each $(c_s, c_o)$ pair, relation distributions are aggregated over time to obtain a stabilized relational prior
$P_{\mathrm{WM}}(r \mid c_s, c_o)$.
During inference, this prior is estimated without accessing ground-truth relations and is updated solely from model predictions.
This temporal accumulation suppresses transient mispredictions and reinforces consistently supported predicates.
In parallel, we compute the visually aggregated relation distribution by summing per-frame logits:
\begin{equation}
P_{\mathrm{vis}}(r \mid i,j)
=
\mathrm{Softmax}\left(
\sum_{t \in \mathcal{T}_{ij}} \ell_t(r \mid i,j)
\right).
\label{eq:9}
\end{equation}
We estimate edge uncertainty via the entropy of $P_{\mathrm{vis}}$.
The world-model prior is injected only when this entropy exceeds a predefined threshold, indicating relational ambiguity.
The final distribution is obtained via uncertainty-aware additive fusion:
\begin{equation}
P_{\mathrm{new}}(r \mid i,j)
\propto
P_{\mathrm{vis}}(r \mid i,j)
+
\alpha_{ij} \, P_{\mathrm{WM}}(r \mid c_s, c_o),
\label{eq:10}
\end{equation}
where $\alpha_{ij}$ is activated only under high uncertainty.
The result is normalized to produce a valid probability distribution.
This design complements missing relation probabilities while avoiding overrides of strong visual and geometric evidence.

\section{Experiments}
\label{sec:exp}

\textbf{Datasets.}
We evaluate DeWorldSG on the 3DSSG \cite{3dssg} and ReplicaSSG \cite{fross} benchmarks. 3DSSG extends 3RScan \cite{3rscan} with 3D semantic scene graph annotations, comprising 1,482 real-world RGB-D scans of 478 indoor environments with instance-level annotations and directional relationships. ReplicaSSG \cite{fross} augments the Replica dataset \cite{replica} with relational annotations and adopts the Visual Genome \cite{vg} object category system to enable zero-shot transfer learning. It uses 7 scenes for validation and 11 scenes for testing.
~\\
\textbf{Evaluation Metrics.}
We follow the evaluation protocol used in prior 3D Semantic Scene Graph work \cite{3dssg, sgfusion, monossg, fross, kim, imp, vgfm}. We report Object Recall, Predicate Recall, Relationship Recall, and mRecall to account for class imbalance. Object Recall measures the fraction of ground-truth object instances that are correctly localized and classified. Predicate Recall measures the fraction of correctly predicted predicates among detected object pairs. Relationship Recall measures the fraction of correctly predicted \textit{subject-predicate-object} triplets for connected object pairs. 
~\\
\textbf{Baselines.}
We compare DeWorldSG against representative existing 3D SSG generation methods. These include methods that explicitly utilize 3D geometry information \cite{3dssg, sgfusion, monossg} and lifting-based methods that first estimate a 2D scene graph and then extend it into 3D space \cite{imp, vgfm, kim, fross}. All methods are evaluated under the same label mapping and evaluation criteria to ensure fair comparison.
~\\
\textbf{Implementation Details.}
All experiments including training, evaluation, and visualization are conducted on a single NVIDIA RTX A6000 GPU with 48 GB of memory. Following \cite{fross}, we apply a confidence threshold of 0.7 for object filtering and retain only the top 10 relationships per frame. In the DR stage (\cref{eq:1}, \cref{eq:2}), we set $\tau = 0.05$, $\epsilon=1e-3$, $\gamma_{\text{base}}$= 0.03, $\alpha$= 10, and $\beta$= 0.02. During global graph merging, we use a Hellinger distance threshold of $\delta_g = 0.7$, and a class distance threshold of $\delta_c = 0.8$. The uncertainty-based relation fusion coefficient is set to $\alpha_{ij} = 0.5$, and the entropy threshold is set to 1.0. For more details, please refer to the Supplementary Material.


\begin{table}[t]
\centering
\caption{Performance comparison of 3D SSG generation methods on the 3DSSG dataset. Note that MonoSSG \cite{monossg} supports RGB inputs, the reported results are based on predictions using dense reconstruction from RGB-D sequences.}
\label{tab:main_results}
\setlength{\tabcolsep}{6pt}
\renewcommand{\arraystretch}{1.1}
\begin{tabular}{l c ccc cc}
\toprule
\multirow{2}{*}{Method} 
& \multirow{2}{*}{Input Modality}
& \multicolumn{3}{c}{\textit{Recall} (\%)} 
& \multicolumn{2}{c}{\textit{mRecall} (\%)} \\
& & Rel. & Obj. & Pred. & Obj. & Pred. \\
\midrule
IMP \cite{imp}      & RGB-D       & 19.7 & 49.5 & 20.9 & 34.7 & 13.8 \\
VGFM \cite{vgfm}    & RGB         & 19.6 & 50.0 & 20.4 & 34.8 & 11.0 \\
3DSSG \cite{3dssg}   & Point Cloud & 12.9 & 37.4 & 22.0 & 26.2 & 14.4 \\
SGFN \cite{sgfusion}     & RGB-D       & 22.0 & 51.6 & 27.5 & 37.7 & 24.0 \\
Kim \cite{kim}      & RGB-D       & 9.1  & 59.0 & 7.1  & 51.0 & 8.0  \\
MonoSSG \cite{monossg} & RGB-D       & 23.3 & 53.8 & 28.4 & 43.8 & 26.6 \\
FROSS \cite{fross}    & RGB-D       & 27.9 & 62.4 & 33.0 & 63.8 & 18.0 \\
\midrule
DeWorldSG (Ours)
& RGB-D 
& \textbf{50.2} 
& \textbf{75.0} 
& \textbf{57.3} 
& \textbf{70.6} 
& \textbf{36.0} \\
\bottomrule
\end{tabular}
\end{table}

\begin{table}[t]
\centering
\caption{Performance comparison using predicted camera trajectories from the ReplicaSSG dataset and ground-truth camera trajectories as inputs. ‘w/o GT Pose’ denotes the DeWorldSG version utilizing trajectories estimated by ORB-SLAM3 \cite{orbslam3}.}
\label{tab:pose_comparison}
\setlength{\tabcolsep}{6pt}
\renewcommand{\arraystretch}{1.15}
\begin{tabular}{l ccc cc}
\toprule
\multirow{2}{*}{Method}
& \multicolumn{3}{c}{\textit{Recall} (\%)}
& \multicolumn{2}{c}{\textit{mRecall} (\%)} \\
& Rel. & Obj. & Pred. & Obj. & Pred. \\
\midrule
FROSS \cite{fross} & 22.3 & 26.1 & 27.8 & 28.8 & 20.4 \\
Ours  & \textbf{38.2} & \textbf{35.4} & \textbf{45.3} & \textbf{38.1} & \textbf{27.3} \\
\midrule
FROSS \cite{fross} w/o GT Pose & 22.7 & 25.8 & 27.2 & 27.7 & 20.1 \\
Ours w/o GT Pose  & \textbf{37.4} & \textbf{33.8} & \textbf{42.8} & \textbf{37.6} & \textbf{26.9} \\
\bottomrule
\end{tabular}
\end{table}

\subsection{Quantitative Results}

We evaluate DeWorldSG on the 3DSSG benchmark, and report the results in \cref{tab:main_results}. 
DeWorldSG outperforms prior state-of-the-art across all metrics, achieving +77.4\% improvement in Relation Recall, +20.2\% improvement in Object Recall, and +23.2\% improvement in Predicate Recall.
Notably, pointcloud-based methods \cite{3dssg, sgfusion, monossg} typically excel at relational reasoning due to explicit 3D geometry. We surpass these methods leveraging only depth-guided lifting and temporal relation accumulation, without scene-level point-cloud inference.
Moreover, under class-balanced mRecall, DeWorldSG demonstrates robust 3D scene graph generation despite severe class imbalance.
On ReplicaSSG \cref{tab:pose_comparison}, DeWorldSG substantially outperforms the baseline model, gaining  +71.3\% in relationship, +35.6\% in Object Recall, and +62.9\% in Predicate Recall.
These results indicate that DeWorldSG’s incremental merging and spatiotemporal relation aggregation strategy across consecutive frames mitigate the isolation of objects and relations that commonly arises in single-frame inference, thereby improving scene-level relational consistency.

\begin{figure*}[t]
\centering
\includegraphics[width=\linewidth]{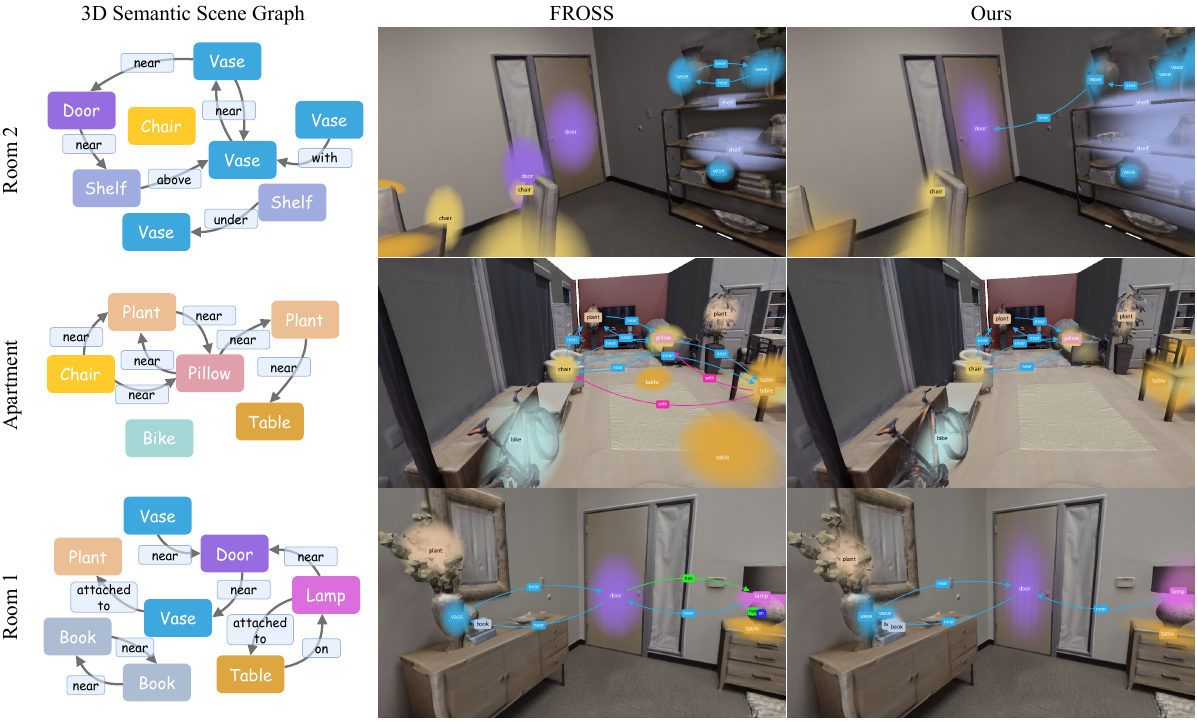}
\caption{
    Qualitative comparison of 3D SSG generation results on ReplicaSSG. (Zoom for details)
}
\label{fig:qualitative}
\end{figure*}

\subsection{Qualitative Results}

\begin{wrapfigure}{r}{0.4\columnwidth}
    \centering
    \vspace{-30pt}
    \includegraphics[width=\linewidth]{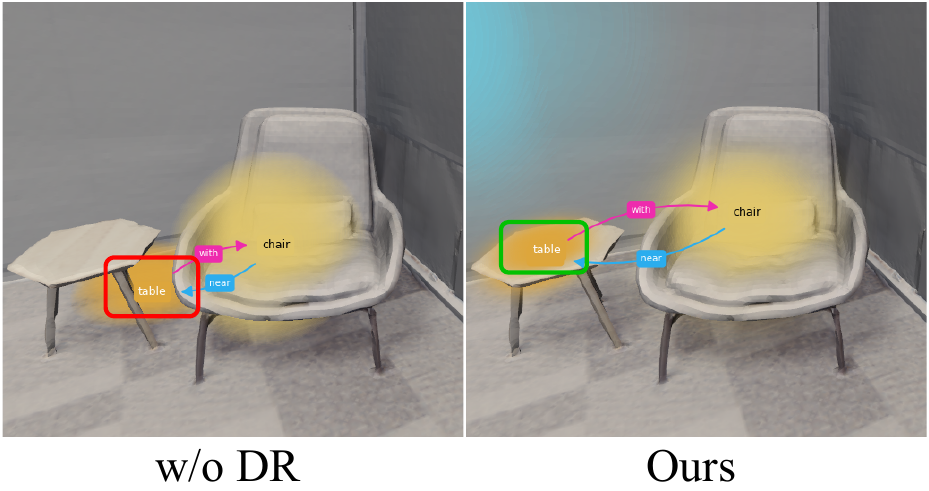}
    \caption{
    3D SSG results under
the same camera viewpoint without (left) and with (right) Dual-Domain Refinement (DR). The red rectangle highlights the unstable object position. 
    }
    \label{fig:ablation_ddr}
    \vspace{-45.5pt}
\end{wrapfigure}

\cref{fig:qualitative} qualitatively compares DeWorldSG's predictions against the previous SoTA method FROSS \cite{fross} on ReplicaSSG.
DeWorldSG produces more consistent object nodes and spatial extents via depth-refined 3D Gaussians (see \cref{fig:ablation_ddr}), and forms semantically coherent relation graphs through temporal evidence accumulation from world models.
In contrast, FROSS frequently exhibits failure cases in which the same object is redundantly instantiated or object locations are inaccurately projected, leading to overly dense or incorrectly connected relation edges.
Additional qualitative comparisons are provided in the Supplementary Material.

\subsection{Ablation Study}

We analyze DeWorldSG's key components through four configurations as presented in \cref{tab:ablation}. 
The baseline model (a), which uses only 2D scene graph lifting without additional refinement, yields the lowest performance across all metrics. Adding box-conditioned instance masking module (b) showed overall performance improvements across all metrics, confirming that mask-guided depth sampling stabilizes 3D Gaussian estimation for both localization and relation reasoning.
Subsequently, the dual-domain depth refinement module (c) further boosts performance by removing depth noise and flying pixels. This validates the reliability of using spatial and depth filtering for global merging.
Configuration (d) replaces the V-JEPA 2 video world-model prior with DINOv2 \cite{dinov2} while keeping the same probe-training and relation-injection pipeline.
This setting isolates the effect of the temporal world-model prior from generic pretrained visual representations.
Although the DINOv2 baseline yields modest predicate gains, DeWorldSG's full temporal accumulation achieves stronger relation performance, demonstrating that spatiotemporal evidence collection provides more informative cues for relation inference than static image-level representations.



To assess camera pose robustness, we replace ground-truth poses with ORB-SLAM3 \cite{orbslam3} estimated poses on ReplicaSSG and report the results in \cref{tab:pose_comparison}.
Our model maintains a strong performance (89.2\% of GT-pose Object Recall and 87.6\% Relationship Recall), confirming stable graph generation under pose noise typical of online AR/robotics systems.

\begin{table}[t]
\centering
\caption{Ablation results on the 3DSSG benchmark. We evaluate four  configurations by progressively enabling instance mask-based depth distribution extraction, Dual-Domain Depth Refinement (DR), predicate refinement with world-model, and DINOv2 \cite{dinov2}-based refinement.}
\label{tab:ablation}
\setlength{\tabcolsep}{5pt}
\renewcommand{\arraystretch}{1.15}
\begin{tabular}{c c c c c ccc cc}
\toprule
\multirow{2}{*}{Model}
& \multirow{2}{*}{\shortstack{Instance\\Mask}}
& \multirow{2}{*}{DR}
& \multirow{2}{*}{\shortstack{World\\Model}}
& \multirow{2}{*}{DINOv2}
& \multicolumn{3}{c}{\textit{Recall} (\%)}
& \multicolumn{2}{c}{\textit{mRecall} (\%)} \\
& & & & & Rel. & Obj. & Pred. & Obj. & Pred. \\
\midrule
(a) & \xmark & \xmark & \xmark & --    & 28.5 & 63.4 & 33.4 & 64.3 & 20.5 \\
(b) & \cmark & \xmark & \xmark & --    & 40.3 & 71.7 & 46.7 & 68.1 & 26.3 \\
(c) & \cmark & \cmark & \xmark & --    & 47.8 & 74.6 & 52.9 & \textbf{70.8} & 30.1 \\
(d) & \cmark & \cmark & --    & \cmark & 49.5& 74.6& 56.6& 69.5& 33.7\\
\midrule
(e) & \cmark & \cmark & \cmark & -- 
& \textbf{50.2} & \textbf{75.0} & \textbf{57.3} & 70.6 & \textbf{36.0} \\
\bottomrule
\end{tabular}
\end{table}

\subsection{Runtime}


We analyze DeWorldSG's runtime in the same environment as \cref{sec:exp}.
The results are averaged over 14,400 frames from four test scenes in the ReplicaSSG dataset to assess practical online feasibility.
Module-specific runtime results show that the object detection and instance processing stage takes an average of 4.42 ms per frame, relation inference 7.28 ms, and global 3D SSG merging 2.23 ms per frame. 
The main computational overhead comes from SAM-based mask segmentation (25.26 ms) and relation refinement process (69.39 ms).
Overall, the full pipeline has an average latency of 108.53 ms per frame, demonstrating practical feasibility for incremental 3D scene graph updates in interactive settings.

\subsection{Limitations}

\begin{wrapfigure}{r}{0.4\columnwidth}
    \centering
    \vspace{-30pt}
    \includegraphics[width=\linewidth]{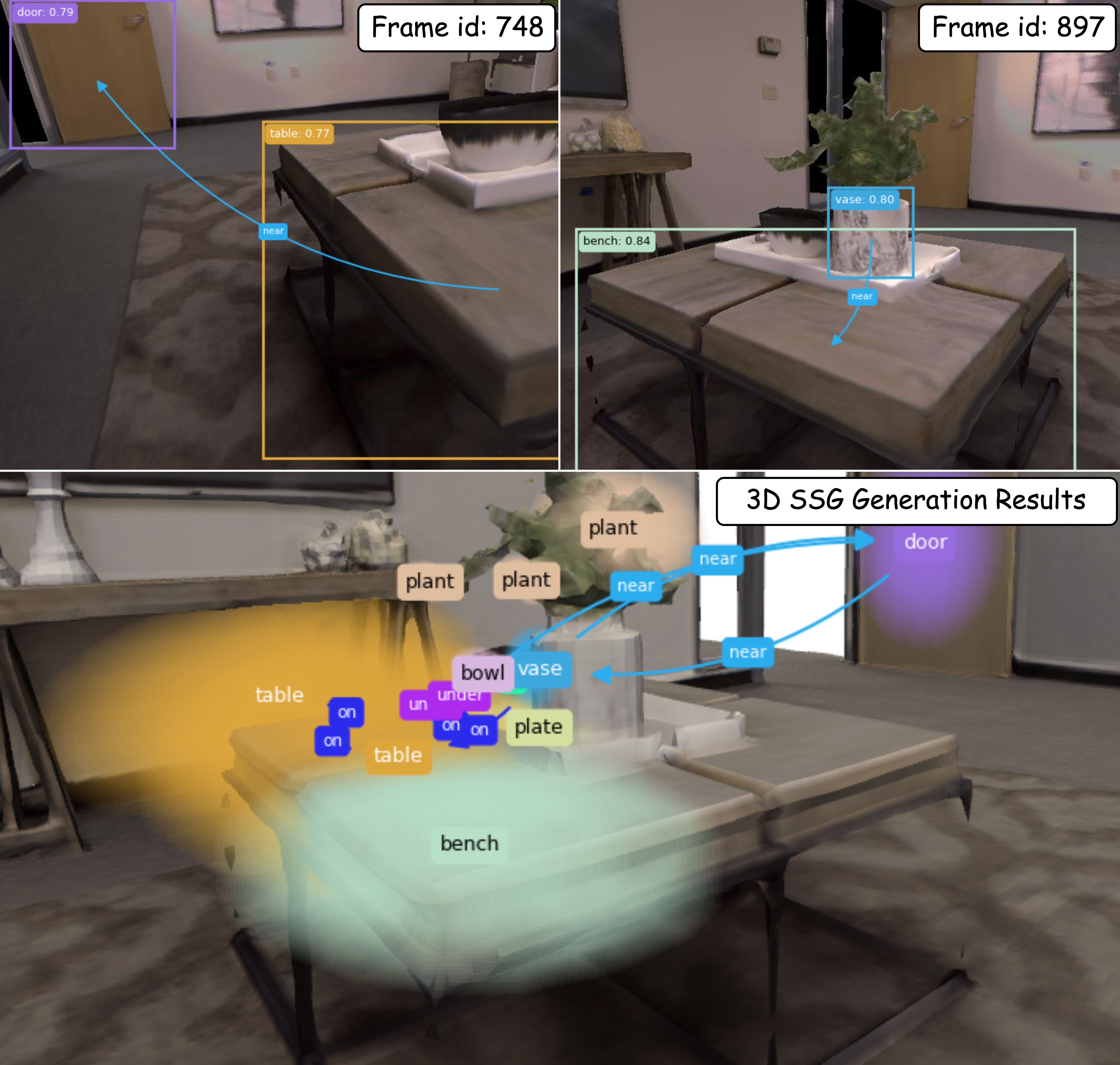}
    \caption{
    Failure case caused by inconsistent 2D object classification across frames.
    }
    \label{fig:fail}
    \vspace{-50pt}
\end{wrapfigure}

DeWorldSG relies on a 2D object detector for object class prediction, which can propagate frame-level recognition errors into the constructed 3D scene graph. As shown in \cref{fig:fail}, when the same physical object (e.g., a table) receives inconsistent labels across frames (e.g., bench), the system may treat them as distinct instances during graph construction, creating duplicated nodes and fragmented relational structures. Such inconsistencies are more likely under partial occlusion or large viewpoint changes. Future work could address this through cross-frame semantic consistency mechanisms or more robust 2D object detectors.

\section{Conclusion}
We presented \textit{DeWorldSG}, a framework for generating spatio-temporally coherent 3D semantic scene graphs from RGB-D sequences. 
By explicitly modeling instance-level depth distributions and leveraging world-model priors for relational reasoning, DeWorldSG enables stable object representation and consistent relation inference under noisy observations. 
Extensive experiments demonstrate state-of-the-art performance on 3D SSG benchmarks and show that our approach maintains robust graph construction even without ground-truth camera poses.

Beyond benchmark improvements, DeWorldSG provides a practical foundation for 3D scene understanding in embodied systems. 
The ability to incrementally construct structured scene representations opens new opportunities for applications such as robotic perception, embodied AI agents, and spatially-aware AR systems that require reliable reasoning over objects and their relationships.
We believe that integrating probabilistic geometric modeling with predictive world-model priors represents a promising direction toward scalable and interactive 3D scene understanding in real-world environments.


\section{Acknowledgements}
I would like to extend my sincere gratitude to all of my co-authors for their invaluable contributions to this manuscript. The figures in this paper were designed using resources from Flaticon.com. This paper was supported by Korea Institute for Advancement of Technology(KIAT) grant funded by the Korea Government(MOTIE) (RS-2025-02304167, HRD Program for Industrial Innovation).
This work was supported by the IITP (Institute of Information \& Communications Technology Planning \& Evaluation)-ITRC (Information Technology Research Center) grant funded by the Korea government(Ministry of Science and ICT)(IITP-2026-RS-2024-00436398). This work was supported by Institute of Information \& Communications Technology Planning \& Evaluation (IITP) grant funded by the Korea government(MSIT) (RS-2026-25523396, Core Technology Development for Immersive Content).

%
%
\bibliographystyle{splncs04}
\bibliography{main}
\end{document}